# Automatic Individual Identification of Patterned Solitary Species Based on Unlabeled Video Data

Vanessa Suessle[1], Mimi Arandjelovic[2,3], Ammie K. Kalan [4], Anthony Agbor[2], Christophe Boesch [2], Gregory Brazzola [2], Tobias Deschner [5], Paula Dieguez [3], Anne-Céline Granjon [2], Hjalmar Kuehl [3,6,7], Anja Landsmann [8], Juan Lapuente [2], Nuria Maldonado [2], Amelia Meier [2], Zuzana Rockaiova [8], Erin G. Wessling [9,10], Roman M. Wittig [11,12], Colleen T. Downs[13], Andreas Weinmann[14], Elke Hergenroether[1],

1. Department of Computer Science, University of Applied Sciences Darmstadt, Darmstadt, Germany
2. Max Planck Institute for Evolutionary Anthropology (MPI EVAN), Leipzig, Germany
3. German Centre for Integrative Biodiversity Research (iDiv) Halle-Jena-Leipzig, Leipzig, Germany
4. Department of Anthropology, University of Victoria, Victoria, Canada
5. Institute of Cognitive Science, University of Osnabrück, Osnabrück, Germany
6. Senckenberg Museum of Natural History Goerlitz, Goerlitz, Germany
7. International Institute Zittau, Technische Universität Dresden, Zittau, Germany
8. Zooniverse Citizen Scientist, c/o Max Plank Institute for Evolutionary Anthropology, Leipzig, Germany
9. Department of Human Evolutionary Biology, Harvard University, Cambridge, Massachusetts, USA
10. School of Psychology & Neuroscience, University of St. Andrews, St. Andrews, Scotland
11. Ape Social Mind Lab, Institute for Cognitive Sciences Marc Jeannerod, UMR 5229 CNRS / University of Lyon 1, Bron, France
12. Taï Chimpanzee Project, Centre Suisse de Recherches Scientifiques, Abidjan 01, Côte d'Ivoire
13. School of Life Sciences, University of KwaZulu-Natal, Scottsville, Pietermaritzburg, South Africa
14. Department of Mathematics, University of Applied Sciences Darmstadt, Darmstadt, Germany

**ABSTRACT**
The manual processing and analysis of videos from camera traps is time-consuming and includes several steps, ranging from the filtering of falsely triggered footage to identifying and re-identifying individuals. In this study, we developed a pipeline to automatically analyze videos from camera traps to identify individuals without requiring manual interaction. This pipeline applies to animal species with uniquely identifiable fur patterns and solitary behavior, such as leopards (*Panthera pardus*). We assumed that the same individual was seen throughout one triggered video sequence. With this assumption, multiple images could be assigned to an individual for the initial database filling without pre-labeling. The pipeline was based on well-established components from computer vision and deep learning, particularly convolutional neural networks (CNNs) and scale-invariant feature transform (SIFT) features. We augmented this basis by implementing additional components to substitute otherwise required human interactions. Based on the similarity between frames from the video material, clusters were formed that represented individuals bypassing the open set problem of the unknown total population. The pipeline was tested on a dataset of leopard videos collected by the Pan African Programme: The Cultured Chimpanzee (PanAf) and achieved a success rate of over 83% for correct matches between previously unknown individuals. The proposed pipeline can become a valuable tool for future conservation projects based on camera trap data, reducing the work of manual analysis for individual identification, when labeled data is unavailable.

## 1. INTRODUCTION

With nearly 40,000 species classified as threatened by the IUCN and a general upward trend [1], efficient and reliable monitoring of wild animals in their natural habitats is essential for wildlife conservation. Monitoring is a complex and time-intensive task for ecologists and is a crucial step to answer hypotheses on the abundance, behavior, territory, social relationships and anthropogenic interaction. Conducting a population monitoring on a species gives scientists insights into the species' endangerment and helps to achieve conservation objectives to protect the population adequately and is an integral part of adaptive conservation cycles [2, 3]. Individual identification is a common method to estimate a population size [4]. Over recent years, camera traps have become an increasingly popular tool to monitor wildlife unobtrusively. The low acquisition and maintenance costs make camera traps an effective tool to collect large volumes of data without invading the habitat and disrupting the animal's natural behavior [5]. The affordability of camera traps generally results in an immense amount of collected videos and images. However, analyzing the enormous amount of data is time consuming, monotonous and exceeds the processing workload experts can manually accomplish in a short time [6].

Computer vision (CV) and artificial intelligence (AI) have the potential to automate selected tasks and support ecologists in their work to identify individuals based on visual characteristics [4]. Convolutional neural networks (CNNs) have the power to learn features and quickly classify images. The drawback of current supervised classification methods is the relatively large amount of required labeled training data, which is not available in most cases for individual identification in wild environments [4]. The pipeline developed for this study, was composed of different components to automate the manual steps typical of individual identification, combining deep learning and classical vision for feature detection as motivated in other studies [7]. The analytical steps included the detection and location of the animal, filtering of empty images and videos, extraction of meta information (e.g. from video files), detection and description of an individual's features, comparing the identified features among individuals and finally, the decision making about potential matches.

We demonstrated the usability of the pipeline with a dataset of leopard (*Panthera pardus*) videos collected with camera traps by the Pan African Programme: The Cultured Chimpanzee (PanAf) [8] (Figure 1). A leopard's coat pattern has the same characteristic as a human fingerprint. Both uniquely identify an individual [9]. We aimed to label and match individuals' appearances in the dataset and assign an ID for each individual, if the available data allowed. This task can be challenging because the data were collected in the wild with varying conditions and differed in terms of lighting, quality, occlusions and included false triggers. Additionally, animals can be hidden, appear from various viewpoints or distances as well as in diverse poses (Figure 2). Leopards fulfill the requirement of being a solitary species [10] and thus an automatic labeling of the data with the presented pipeline is applicable. We therefore could reasonably assume that within one motion-triggered video, the same individual was seen throughout the frame sequence, which enabled the collection of different footage of the individual. A general drawback of CNNs for classification tasks is the open set problem [11, 12]. A traditional classifier can only re-identify and sort into a dictionary of known classes it was trained on. The classifier is compelled to pick the class that fits the most, even if none of the classes fit from a human perspective. For individual identification, this means that for an unknown individual, the classifier assigns it to one of the known individuals that fits the most. For population monitoring studies the total population is not known in advance and identifying unknown individuals is of high relevance. Our aim was to assist ecologists with a tool for individual identification of fur-patterned solitary species without requiring a large, labeled dataset of known individuals and the necessity of constant user inputs. We developed a modular pipeline that covers the subtasks of data preprocessing to individual identification.

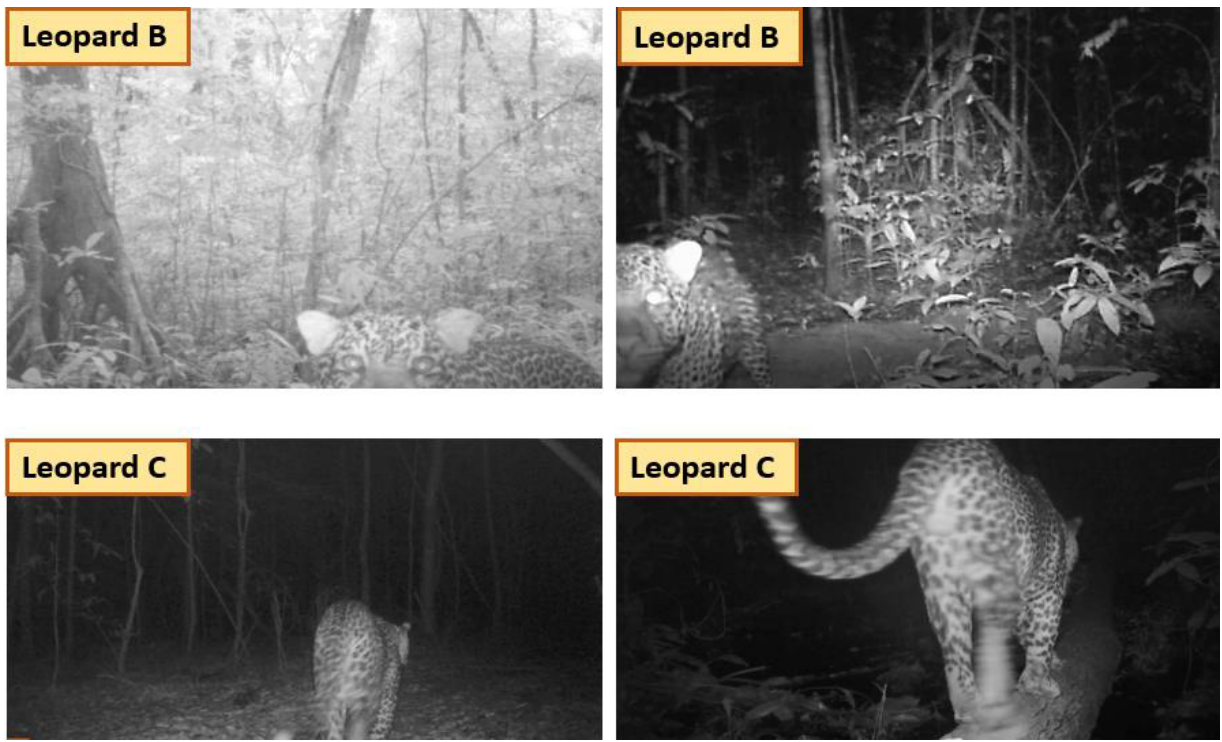

**Figure 1. Examples of correct matches of individuals from different videos of low illumination and quality, with only parts of the animals being visible. The individuals in the images on the left were matched to the individuals in the images on the right respectively.**

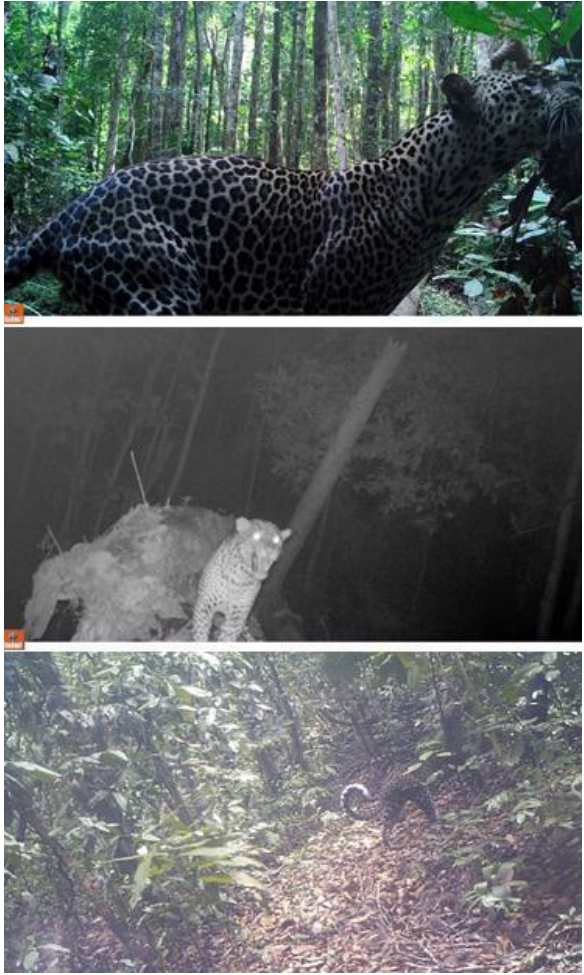

**Figure 2. Footage of leopards from camera traps from the PanAf with different quality, lighting, visibility and posture of the animal.**

## 2. RELATED WORK ON CAMERA TRAP DATA ANALYSES

### Detection & Classification

Non-prefiltered datasets taken by motion-triggered cameras usually include a relatively large amount of falsely triggered images or videos, and footage on a spectrum of many species living in the ecosystem. An automated detector and classifier are essential for ecologists to process the automatically captured data, in a reasonable time frame [13]. For both filtering tasks, a type of classification is needed, either for the species or more generally separating into 'non-empty' and 'empty' classes. For the detection and localization of wildlife, which covers the task of filtering empty images, the MegaDetector [14] model is state-of-the-art. The trained CNN model returns bounding boxes around the detected animals. It was trained on many different datasets, including different species taken in diverse ecosystems. The model is constantly improved, and new versions are released regularly. On an ordinary GPU, the MegaDetector can process between 150,000 and 250,000 images per day [14].

While the MegaDetector is applicable to ecosystems around the globe, species classification approaches are usually tied to a specific ecosystem and its inhabitants. Pre-trained CNN models for species in North America [15, 16], Africa [16, 13, 17], Europe [18, 19] and Australia [16] are available, but they mostly cover focal species. Trained CNN models are also prone to the open set problem and can only classify the species they were trained on and are not sensitive to unknown species. Training such models require a large amount of labeled training data, and manual labeling is time-consuming [20]. For datasets with thousands or even millions of records, the labeling of the data may last multiple years [21]. To speed up this process, platforms were created to involve volunteers labeling the data.

### Citizen Science

Volunteers who label data for projects are called citizen scientists [13, 22]. Platforms like Zooniverse [23], Wildlife Insights [24] and Wildbook [25] offer an option for research projects to open their data to citizen scientists who sort the images into predefined classes. With this approach, organizations can process the data relatively faster, and with the positive side effect of drawing the public's attention to wildlife conservation. Besides the progress for the current case study, labeling the datasets benefits the training of machine learning algorithms to support future wildlife conservation projects. The Snapshot Safari project [6, 26] is one of the world's largest camera trapping initiatives that used citizen scientists. From 2013 to 2020, over 138,000 volunteers from across the globe labeled more than nine million images. The drawback of this approach is that volunteers usually do not have many years of expertise and lack the knowledge to label rare species or individuals which can be challenging with camera trap images, even for experts [27]. The first attempts for online data processing with citizen scientists in near-real-time were conducted by a project in South Africa to fight wildlife poaching [28]. Captured images were immediately uploaded to a website. Volunteers examining the data can report a suspected poaching vehicle or human in the images and trigger a warning to local rangers in the nature reserve who thereby gain the opportunity to react quickly and prevent poaching activities.

## Identification and Re-Identification of Individuals

While the automated classification of different species has been investigated over recent years, the field of identification of individual animals is still in its infancy. The identification and re-identification of individuals differ from the above-described task of species classification. For this task, not the species is recognized, but the unique individual. The identification of individuals with computer vision methods relies on visual biometric features that uniquely identify the individual. Since the biometric characteristics of different species vary, no overall solution covers all case studies. Early computer vision-assisted approaches for individual identification of marine mammals used unique body marks on the fins [29], or the fin's trailing edge was represented as integral curvatures [30]. The first ever automated estimation of a population was performed on African penguins (*Spheniscus demersus*) based on spot locations on their chest [31] compared against a database of known individuals. Nowadays, CNNs are a popular solution for individual identification and re-identification tasks. CNNs can extract features from animals with distinctive body marks. Previous studies applied CNNs to coat-, skin- or feather patterns of ringed seals (*Pusa hispida*) [32], whales [33], snow leopards (*Panthera uncia*) [34] and small birds [35]. In a study on the Great Barrier Reef, shell patterns of green turtles (*Chelonia mydas*) were extracted with a neural network system [36].

To the best of our knowledge, only supervised learning models have been used for individual identification of wildlife, treating each individual as a single class. Supervised, CNN-based solutions require large training datasets of labeled images, which are usually not available for wild animals, especially for automatically captured data.

A group from Shanghai Jiao Tong University, together with the World Wide Fund for Nature (WWF), generated and published a labeled dataset of 92 Amur tigers (*Panthera tigris altaica*) for training purposes. They trained an individual identification model on this dataset, for which each flank of a tiger was treated as an entity [37]. The patterned pelages on opposite flanks of felids are different and independent [38]. When the footage only shows opposite flanks of an animal separately in different captures, with no overlap of body parts seen, it is impossible to recognize whether the flanks belong to the same individual. Treating both flanks of the same individual as separate entities could lead to a biased estimation of the population size by a factor of 2. Further research tested Siamese convolutional neural networks with triplet loss [39], which are commonly used for person re-identification [40], for the re-identification of lions (*Panthera leo*), nyalas (*Tragelaphus angasii*) and ringed seals [41, 42]. But as with other deep learning approaches, labeled training data are required. Furthermore, CNN-based solutions bear the open set problem, which complicates identifying entities unknown to the population.

An alternative to CNN-based solutions is the pattern-matching scale-invariant feature transform (SIFT) algorithm [43]; with its scale, location, viewpoint and illumination invariant feature descriptor, it is well-suited for camera trap data [44]. Wild-ID [45] and HotSpotter [46, 25, 47, 48] are individual identification programs based on the SIFT algorithm for species with distinctive visual features. The SIFT approach is not applicable for species that lack unique fur or body markings.

The same concept used for human facial recognition can be applied to identify primates [49–51], pandas [52], bears [53] and pigs [54, 55]. For approaches concentrating on the face the collection of useable datasets in terms of quality, viewpoint and labeling is even more difficult than for fur-patterned species. Available datasets mostly stem from captive animals from zoos or farms.

The proposed solutions for individual identification described above all had at least one of the following: a labeled dataset, data collected under non-wild conditions, images manually photographed, closed populations where all individuals were known, or the solutions required human decision making or drawing bounding boxes.

In contrast, our study presents a pipeline that does not rely on labeled data or human interaction and covers the open set challenge. Our pipeline was tested with a dataset of videos automatically captured with camera traps in the wild. Thus, it covers the task of individual identification and re-identification for an unlabeled dataset. Our pipeline benefits research by saving users valuable time estimating the number of individuals in a dataset.

## 3. PIPELINE FOR AUTOMATIC INDIVIDUAL IDENTIFICATION

Our objective was to provide researchers with a tool that automates individual identification from determining the animal's location in the video frames to feature extraction and matching. We develop a robust pipeline that unites the aforementioned analytical steps. The pipeline consists of newly developed components combined with existing components with proven functionality in prior case studies. Interim steps are implemented to substitute the otherwise required user input for specific components. The pipeline's main components cover the following tasks:

1. Image extraction: Extracting frames from video files and incorporating additional sequence-based information.
2. Object detection: Locating the animal within the image or classifying an image as empty.
3. Species classification: Selecting only images that include the species of interest.
4. Feature extraction: Detecting and describing features and measuring similarity to other images based on distance.
5. Clustering: Automatic matching of images to individuals based on their similarity.

The components of the pipeline are schematically outlined (Figure 4) and described in more detail below.

### Image Extraction

The first component of the pipeline was the extraction of frames and additional information from video data compared with image data. We assumed that within one triggered video the same individual was seen throughout the frame sequence as leopards lead mainly solitary lives, and multiple images could be initially assigned to one individual ID in the database. Ideally, the animal moved during the video and images of various body poses from different viewpoints were obtained.

### Object Detection

We employed the MegaDetector [14] described above for object detection, which located the animal in the image and returned a bounding box. If no animal was found, the image was classified as empty.

### Species Classification

Depending on the project/species of interest, a specific classification model must be chosen, e.g. the Zamba Cloud [17]. (For potential options, refer to the Related Work section). The present work focuses on individual identification, and the used dataset was already prefiltered for leopards by citizen scientists on the Zooniverse platform [23] the species classification component is therefore greyed out in Figure 4.

### Feature Extraction

For feature detection and feature description, we employed the SIFT-based HotSpotter [46, 25, 47, 48]. SIFT-based algorithms do not require labeled training data. The HotSpotter outperformed its competitor Wild-ID in other studies [44]. The SIFT algorithm identified stable points in the image. It detected and described distinctive and characteristic features of the individual's fur-pattern (Figure 3) and turned them into feature vectors, which were mapped into a feature space. The feature vectors from frames from different videos were queried against other frames in the database, and a similarity score is calculated. The similarity score depended on the Euclidean distance of the mapped feature vectors in the vector space for each frame pairing. The analysis process required distinctive visual features and was only applicable to species with visually distinctive characteristics.

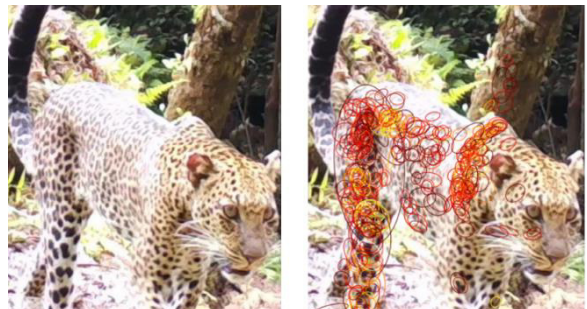

**Figure 3. Left: raw frame from camera trap captured video. Right: extracted SIFT features visualized with HotSpotter [46, 25, 47, 48].**

### Clustering

In the last step, the footage is assigned to respective individuals. This usually requires the user's decision and input. For our presented pipeline, the matching step was conducted automatically. We merged videos into clusters based on the user's predefined threshold for the similarity score. The clusters could be visualized in graphs, where nodes represented videos. Two nodes were connected with an edge, if frames from the videos matched. Each cluster represented one individual. We derived the width of the edges in the visualization (Figure 5) from the degree of similarity. A wide edge implied a high similarity between the animal shown in the frames of the videos. The distance between the nodes and clusters did not give information on their similarity and were arranged to display a comprehensible representation. For each compared video pairing, three cases were possible:

A. Both nodes did not belong to a cluster yet.
B. One node was already part of a cluster, but the other one was not.
C. Both nodes already belonged to a cluster, but different clusters, causing a conflict.

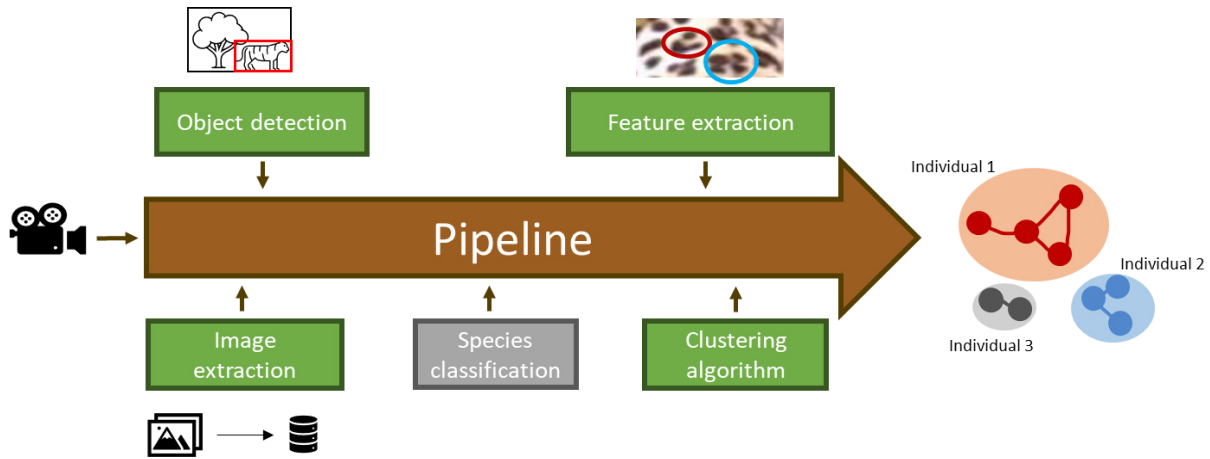

**Figure 4. Schematic concept of the pipeline with its components, including image extraction, object detector, species classifier, feature extractor and matching, finalized by clustering the videos to represent individuals. The species classifier is greyed out and was not used in this case study.**

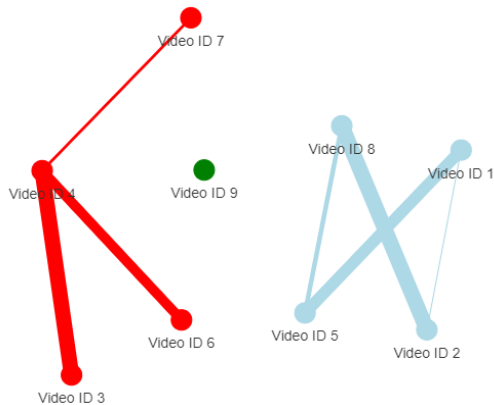

**Figure 5. Schematic illustration of an exemplary cluster output. Each node represents one video and each cluster an individual. The red graph showed that the animals seen in videos 3, 4, 6 and 7 were likely the same individual. Analogue for the blue graph. The animal in video 9 was not matched to any other animal in the video footage.**

We handled the three cases as follows. In case A, a new cluster consisting of the two videos was created, while in case B, the free node was assigned to the existing cluster. Case C is the most complex of the cases. If the compared nodes were already assigned to different clusters, the nodes were rearranged and assigned to another cluster, so that the edges were based on the highest similarity scores. If the similarity of the new video pairing was higher than the similarity that binds the video into the present cluster, the video was released from the present cluster by deleting the edge and creating a new edge to the video with the higher similarity. Figure 5 illustrates a schematic example where the red subgraph shows that the animal seen in videos 3, 4, 6 and 7 were likely the same individual. The same accounts for the animal seen in the videos of the blue subgraph. While for video 9, printed in green, no match was found. Our pipeline outputted an HTML visualization of the clusters and a database comprising similarities and affiliations that ecologists could use for further observations.

The outlined components of the pipeline, including image extraction, object detection, species classification, feature extraction and clustering, are schematically outlined in Figure 4.

### 4. CASE STUDY

The data we used to demonstrate the pipeline in this study were provided by the PanAf [8]. The PanAf collected data at over 40 temporary and collaborative research sites with motion and infrared-equipped camera traps across Central and West Africa. Over 600,000 video clips were taken with a duration of one minute each. Forest habitats are complex areas to collect images and video data. Low light levels during the night further complicate the data collection and analysis, and the videos can be of low quality and only black and white. Snapshots taken from the videos of the animals can vary in distance and be blurred or relatively close up (Figure 1). The PanAf study was originally designed to capture data on chimpanzees and the camera locations were selected to suit their behavior, which makes the dataset especially challenging for other species than chimpanzees. To demonstrate the pipeline, a subset of the leopard dataset was used for which volunteers on Zooniverse labeled the individuals. The leopards' IDs were confirmed when citizen scientists that have been extensively involved and experienced in leopard identification unanimously agreed on the matching spot patterns after manual visual inspection [56]. The information on the individuals was only used for validation purposes and not in the process itself. The leopard subset encompassed footage from 2011 to 2018 and totaled 210 videos from eight field sites representing 68 unique camera locations.

## 5. RESULTS

We demonstrated the pipeline for the individual identification and re-identification for an unlabeled dataset without manual interaction using part of the PanAf leopard dataset.

We automatically processed the 210 videos to validate our pipeline. A total of 116 matches were found, with 97 of those matches being correct, giving 83.6% success rate. The threshold for the similarity score to define a match can be selected to meet the individual project need depending on the use. Matching with a lower threshold allowed more potential matches, giving a chance to match more individuals, even if the underlying footage was challenging, but increased the risk of false matches. A higher threshold requires a higher similarity for a match and led to a lower total value of incorrectly matched individuals, but bore the risk of missing to match captures of the same individual. Depending on the specific use case, preferences must be defined and the threshold must be set accordingly. The image in Figure 6 shows a correct match. In this example, the leopard's visible right hind limb had the most prominent features.

Even for complex footage at nighttime, with low quality and only parts of the animals captured, matching features could be extracted and matched (Figure 1). The most frequent reason for mismatches was the background for images taken at the same location since camera traps were fixed to a site and scenery. With a predefined high threshold for similarity, 100% of the mismatches were caused by matching objects in the background (Figure 7).

## 6. DISCUSSION & OUTLOOK

In this study, we addressed the problem of animal re-identification from camera trap data. Our aim was to substitute manual user input and the need for labeled data. The core idea of the developed pipeline was to take advantage of video data and its consecutive frames for animals with solitary behavior. The pipeline's functionality was proven by identifying and re-identifying leopards from an unlabeled dataset collected by the PanAf.

For future work, a more detailed localization and extraction of the animal from the background rather than the current rectangular bounding boxes can address the challenge of matching the same objects in the background because of the fixed scenery in camera traps. The fixed scenery can also be used as an advantage for the extraction of the background. The detection of objects of interest may also be supported through the availability of video data by extracting optical flows [57]. CNN-based approaches for semantic segmentation could extract a mask for the animal, which excludes the background [58, 59].

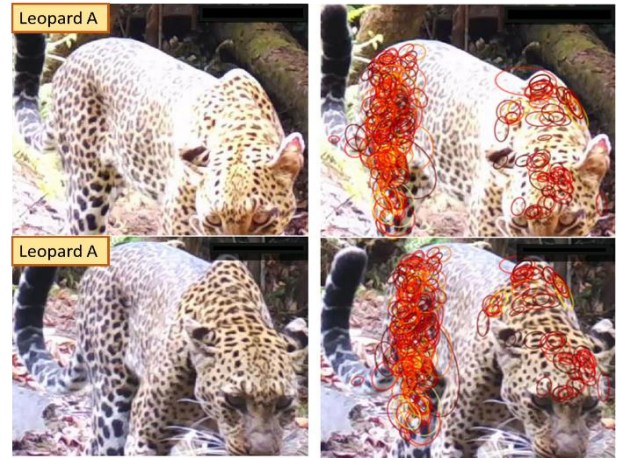

**Figure 6.** An example for a correct match of an individual captured in frames of different videos. The top and bottom row show the same individual in different captures. Left: raw images. Right: the same image, but with detected and matched features in the other image of the same individual.

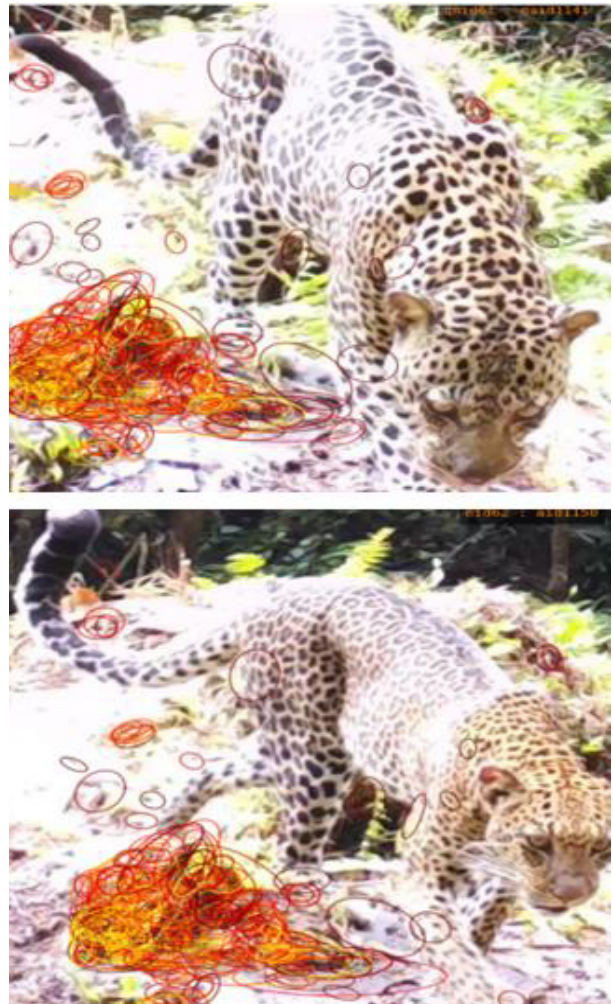

**Figure 7.** Incorrect matching of individuals because of matching objects in the background of fixed camera sceneries.

For future studies, it may be interesting to collect additional information on the viewpoint and the visible flank of the animal for a better overview on known individuals. A process that automatically feeds a database with this information can further improve the monitoring of wildlife and prevent the incorrect matching of opposite sides of animals. With this information not only matching pairings can be detected, but also pairings that reliably show different individuals can be detected and marked.

Our pipeline will be used to support the PanAf identifying additional individuals in other regions. Another experiment for the future is to apply the pipeline to other species and valuate its suitability for cross-species applications.

## 7. ACKNOWLEDGMENTS

We thank the following members of the PanAf consortium and team for data or technical support with data processing: Karsten Dierks, Dervla Dowd, Henk Eshuis, Theo Freeman, John Hart, Thurston Cleveland Hicks, Jessica Junker, Vincent Lapeyre, Vera Leinert, Yasmin Moebius, Mizuki Murai, Emmanuelle Normand, Colleen Stephens and Virginie Vergnes. Further, we thank the following citizen scientists for their support: Tonnie Cummings, Carol Elkins, Lucia Hacker, Briana Harder, Karen Harvey, Laura K. Lynn, Heidi Pfund, Kristeena Sigler, Libby Smith, Jane Widness and Heike Wilken.

We thank the following government agencies for their support in conducting field research in their countries: Ministere des Eaux et Forets, Côte d'Ivoire; Ministère de l'Enseignement Supérieur et de la Recherche Scientifique, Côte d'Ivoire; Institut Congolais pour la Conservation de la Nature, DR-Congo; Ministere de la Recherche Scientifique, DR-Congo; Agence Nationale des Parcs Nationaux, Gabon; Centre National de la Recherche Scientifique (CENAREST), Gabon; Société Equatoriale d'Exploitation Forestière (SEEF), Gabon; Forestry Development Authority, Liberia; Direction des Eaux, Forêts et Chasses, Senegal. As well as the following NGOs in their countries: Taï Chimpanzee Project, Côte d'Ivoire; Wild Chimpanzee Foundation, Côte d'Ivoire, Liberia & Guinea; Lukuru Wildlife Research Foundation, DR-Congo; Loango Ape Project, Gabon; Fongoli Savanna Chimpanzee Project, Senegal.

The Pan African Program: The Cultured Chimpanzee is generously funded by the Max Planck Society, the Max Planck Society Innovation Fund, and the Heinz L. Krekeler Foundation.